\documentclass[fleqn,10pt]{wlscirep}
\usepackage[utf8]{inputenc}
\usepackage[T1]{fontenc}
\usepackage{graphicx} 
\usepackage{lscape}

\title{Building Machine Learning Challenges for Anomaly Detection in Science}

\author[3]{Editors:~Elizabeth G. Campolongo}
\author[2]{Yuan-Tang Chou}
\author[1]{Ekaterina Govorkova}
\author[10]{Wahid Bhimji}
\author[3]{Wei-Lun Chao}
\author[10]{Chris Harris}
\author[2]{Shih-Chieh Hsu}
\author[4]{Hilmar Lapp}
\author[6]{Mark S. Neubauer}
\author[9]{Josephine Namayanja}
% ...

\affil[1]{MIT, Cambridge, MA 02139, USA}
\affil[2]{University of Washington, Seattle, WA 98109, USA}
\affil[3]{The Ohio State University, Columbus, OH 43210, USA}
\affil[4]{Duke University, Durham, NC 27708, USA}
% ...
\affil[6]{University of Illinois Urbana-Champaign, Urbana, IL 61801, USA}
% ...
\affil[9]{University of Maryland Baltimore County, Baltimore, MD 21250 USA}
\affil[10]{Lawrence Berkeley National Laboratory, Berkeley, CA 94720 USA}
% ...

\begin{abstract}
Scientific discoveries are often made by finding a pattern or object that was not predicted by the known rules of science. Oftentimes, these anomalous events or objects that do not conform to the norms are an indication that the rules of science governing the data are incomplete, and something new needs to be present to explain these unexpected outliers. Finding anomalies can be confounding since it requires codifying a complete knowledge of the known scientific behaviors and then projecting these known behaviors on the data to look for deviations. When utilizing machine learning, this presents a particular challenge since we require that the model not only understands scientific data perfectly but also recognizes when the data is inconsistent and outside the scope of its trained behavior. In this paper, we present three datasets aimed at developing machine learning-based anomaly detection for disparate scientific domains covering astrophysics, genomics, and polar science. We provide a scheme to make machine learning challenges around the three datasets \textbf{F}indable, \textbf{A}ccessible, \textbf{I}nteroperable, and \textbf{R}eusable (FAIR). Furthermore, we present an approach that generalizes to future machine learning challenges, enabling the possibility of large, more compute-intensive challenges that can ultimately lead to scientific discovery. 
\end{abstract}

\begin{document}

\maketitle

\section{Introduction}
\label{sec:intro}
Anomaly detection, the process of finding an outlier in a distribution, has long been a critical element of scientific discovery. New anomalies often trigger the question of how to construct a broadly inclusive model allowing for a deeper understanding of the data and, ultimately, a better scientific model. In some cases, this has happened with the observation of a new event or process. Other times, discoveries have been made through aggregating data to observe unexplainable trends~\cite{Pang_2021,boniol2024divetimeseriesanomalydetection,blázquezgarcía2020reviewoutlieranomalydetectiontime,Krenn_2022}.

In this paper, we present a series of machine learning~(ML) challenges targeting anomaly detection in different scientific domains. We define this as the ability to identify events, objects, or phenomena that are rare and do not match ordinary detector noise or otherwise known signals or phenomena. Anomaly detection is thus searching for the “unknown,” including “known-unknowns” (hypothesized and anticipated but otherwise poorly modeled or quantified) and “unknown-unknowns” (not accommodated by prior knowledge in any given domain that---if detected---push the frontier in understanding the theory and phenomenology of the underlying observations and measurements). This could be the identification of 
poorly modeled phenomena, such as a core-collapse supernova
signal in a gravitational-wave detector\cite{Raikman:2023ktu,raikman2024neuralnetworkbasedsearchunmodeled,O2magnetarbursts,Sutton_2010,PhysRevD.95.104046}, unexplained sea ice retreat \cite{devnath2024cmad}, extreme snow melts in the Arctic \cite{10640794}, 
or a new species of animal~\cite{souza2021new,ito2022phenotype,pastore2020annotation}. 
The challenges presented here aim to bring awareness to the difficulty of using ML for these types of problems: the options of well-labeled datasets or even the possibility of training through a systematic procedure, such as token masking, are often insufficient to answer this type of question. Despite that, there is tremendous potential for algorithms focused on anomaly detection to enable scientific discovery. 

The process of using ML 
for anomaly detection, sometimes referred to as outlier detection, has  a long history~\cite{hendrycks2018deep,schlegl2017unsupervised,zong2018deep,yang2024generalized}. 
 However, success has been limited due to the ambiguity of quantifying scientifically meaningful deviation~\cite{galil2023framework,salehi2021unified}.
Nevertheless, several approaches have been tried~\cite{salehi2021unified,yang2024generalized,2025MNRAS.537..931D}, including the use of unsupervised algorithms~\cite{Cui_2023}, which aim to build a mapping for what is regular through an embedding in a latent space, semi-supervised algorithms, which aim to use partially labeled data to construct representation space characteristic to the problem at hand~\cite{Kuusela_2012}, and weakly-supervised algorithms, which aim to play regions of data off of each other to find the largest incongruous regions of the data\cite{jiang2023weaklysupervisedanomalydetection}. 

The ML challenge project brings together communities involved as part of the NSF Harnessing the Data Revolution (HDR) program, dedicated to connecting domain science with AI to enable groundbreaking scientific discoveries.  All five HDR institutes were engaged in developing datasets with a tractable problem defined across their domains, each targeting a single anomaly detection problem and dataset. Ultimately, we narrowed the challenge to three datasets to ensure that they were sufficiently aligned that the same algorithmic toolkit was applicable to each of the challenges. Lastly, the most critical component of this challenge was the ability to hide pre-identified anomalies in such a way that there was no ambiguity in the anomaly definition in the final dataset. The challenge is to see if the proponents can find these pre-hidden anomalies with the smallest possible false positive rate.

The three challenges cover a wide variety of data types and problems, ranging from  time series data, to photographic images, and satellite imagery over time, as described below. 
\begin{itemize}
\item \textbf{Anomalous Gravitational-Wave Transients: } Short-duration (O(1s)) 
astrophysical signals present in public data from the two LIGO gravitational-wave detectors. 
The data that is received from each of the LIGO interferometers is a gravitational-wave strain from a signal, indicating a fractional stretching and squeezing of space-time on the order of $10^{-22}$ the measured length scale. 
The challenge is to find a signal that appears in both detectors and has broad features representative of an astrophysical signature. Such a signature may correspond, for example, to a core-collapse supernova, a white noise burst of unknown origin, or a black hole merger beyond our nominal modeling and expectation from general relativity. 
\item \textbf{Butterfly Hybrids: } Images of butterflies from two mimetic species, each with highly varied subspecies~\cite{heliconius2012butterfly}. It includes images of the various subspecies, as well as different “hybrids” among each species' subspecies. The challenge is to find the anomalous butterfly hybrids without confusing them for subspecies of the other species. 
\item \textbf{Anomalous Sea Level Rise: } Satellite imagery \cite{CDS_portfolio} depicting observed sea levels over time. The challenge is to use this imagery to identify anomalous sea level rise at ports on the east coast of the US. 
\end{itemize}

\begin{figure}[tb]
\centering
\includegraphics[width=0.7\textwidth]{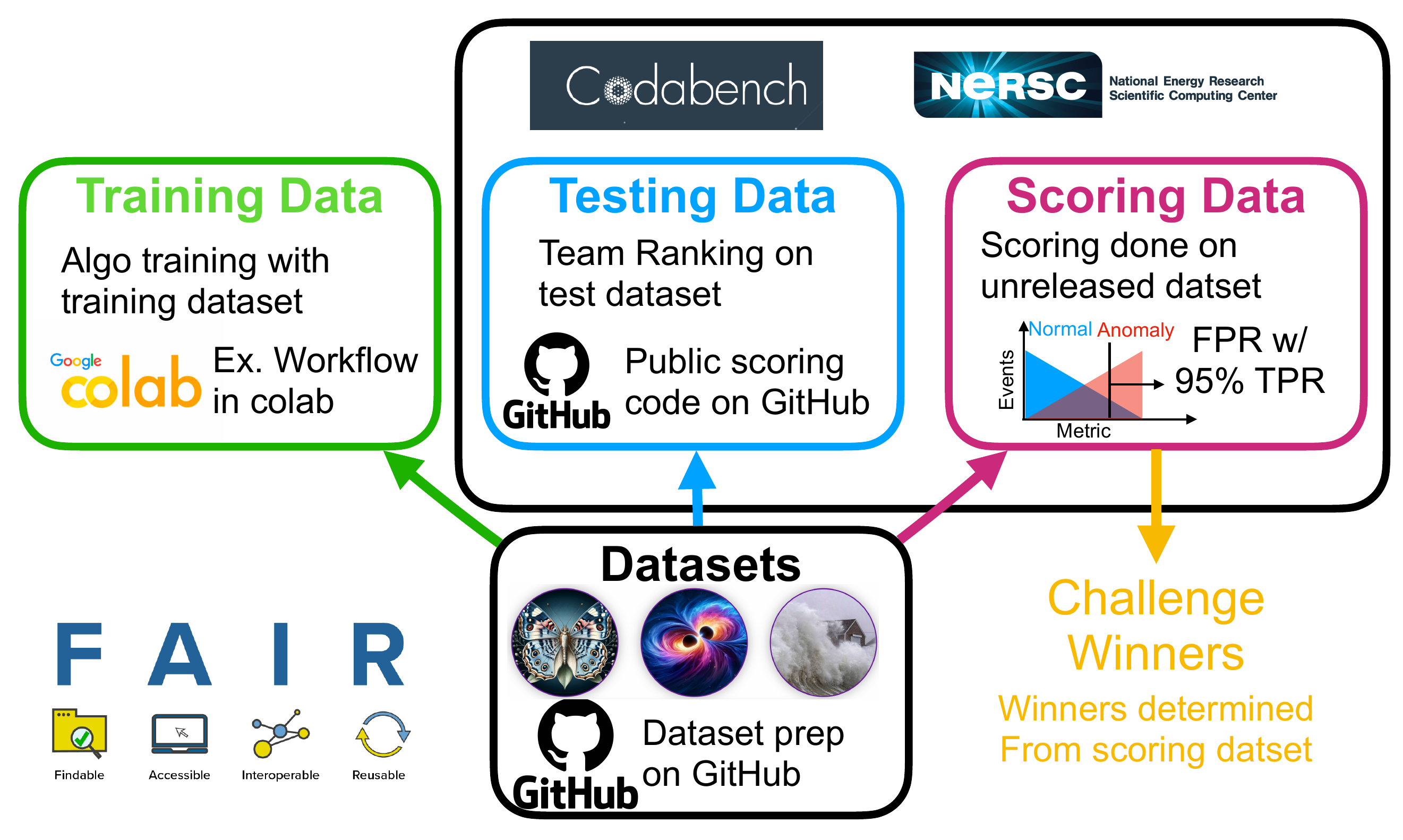}
\caption{Overview of the flow of the ML challenge, and the datasets used for training. All preparation was explicitly made following FAIR principles, with the curation and data cleaning workflows stored on GitHub. The Codabench platform and NERSC computing cluster are used for scoring and ranking of submissions.}
\label{fig:challengeflow}
\end{figure}

A diagram of the challenge flow is provided in Figure~\ref{fig:challengeflow}. For the challenge, teams must develop an algorithm that processes the data and outputs an anomaly score. The scoring is then determined by seeing if the algorithm is capable of singling out the anomalous features of the dataset. Each challenge has three core data subsets: 
\begin{itemize}
    \item \textbf{Training Dataset:} 
    A publicly released subset consisting of the background (non-anomalous) cases and sample anomalous signals to enable the model construction. These signals do not 
    encompass all variations of the challenge dataset, instead providing examples of what \emph{might} be present. Moreover, the background dataset need not have the same proportional composition as the challenge dataset.
    \item \textbf{Testing Dataset:} 
    During the competition, this subset is used for benchmarking algorithms and presenting a leaderboard on the competition platform. It largely consists of background non-anomalous events with some hidden anomalies (known-unknowns identifiable through additional information not available to participants). This dataset is not released publicly; however, participants may run their algorithm on it through the collaboration platform. 
    \item \textbf{Challenge Dataset:} This subset is used to determine the winner. It is constructed in the same way as the testing dataset, but is not used before the final scoring. Thus, no participating team can overly tune the algorithm on the dataset. 
\end{itemize}


A driving goal of this work is to codify a list of best practices that ensure workflows, no matter the scientific domain, are easy to follow and reproduce. Through this work, we endeavor to establish an exemplar for future ML challenges, 
preserving the level of Findability, Accessibilty, Interoperability, and Reusability (FAIR) present in this challenge in future challenges. Moreover, by ensuring a high level of FAIRness we aim to use these datasets for future benchmarking\cite{galil2023framework,Fair4AIWorkshop,zhao2024taxonomychallengescuratingfair}. 

The ensuing paper presents the details of the challenge. In Section~\ref{sec:motivation}, we outline the motivation for the challenge and the scientific problem of anomaly detection. Subsection~\ref{sec:FAIR} highlights 
how the challenge was made FAIR. 
We then present the various datasets comprising the greater challenge in Section~\ref{sec:challenges}, followed by a detailed description of the evaluation metric in Section~\ref{sec:metrics}, and we end with the conclusions.

\section{Scientific and FAIR Motivation}
\label{sec:motivation}
The anomaly detection challenges presented in this paper build on the traditional anomaly detection by providing unique, scientific elements that distinguish them from previous challenges. With the genomics problem, we utilize subtle features of images to replace the need for tedious DNA analysis to differentiate various species and subspecies. With the gravitational-wave challenge, we aim to codify all allowed laws of astrophysics within a time series to enhance the identification of unknown anomalies. With the climate science problem, we replace dedicated sensor data with large-scale, coarse satellite imagery and utilize this to find deviations from the norm. More conventional anomaly detection simply targets outliers; with each challenge here, we add a scientific element that pushes the boundaries of conventional approaches. The challenge of codifying scientific elements within ML adds an additional layer of complexity and highlights the uniqueness of this work.

For each component of this challenge, we constructed anomalies in a controlled manner, utilizing scientific knowledge or data not provided in the challenge. With the genomics problem, this involved using DNA. 
For the climate science problem, we utilized local water level sensors. For the astrophysics problem, knowledge of astrophysical phenomena is used to simulate hypothetical anomalous signals not observed within the data.
With each of these challenges, our use of hidden labels, sensors and simulations is critical to the dataset construction. However, in real-world scenarios this additional information is not present, nor easily obtained. As a result, devising a strategy that can rely on the limited information---even if imperfect---would lead to impactful scientific results applicable to a large amount of unexplored data. 

The choice of these datasets or tasks can be viewed as a seed to more complex problems. By constructing a FAIR challenge, with well-defined scoring metric and clear datasets, we aim to understand effective scientific approaches to anomaly detection. Future challenges can then be adapted to explore more difficult problems, and ultimately to true problems in the scientific domain where solutions might not yet exist at all. 


\subsection{FAIR}
\label{sec:FAIR}
We endeavor to create an entirely \textbf{F}indable \textbf{A}ccessible \textbf{I}nteroperable \textbf{R}eusable, and ultimately reproducible, ML challenge. This choice of FAIRness extends beyond conventional FAIR datasets to include all aspects of the ML challenge. To this end, we require that all components---the challenge code, software environment, datasets, and metadata---adhere to the FAIR principles~\cite{fairguiding} and utilize methods described in ~\cite{Chen_2022,Duarte:2022job,Fair4AIWorkshop}. Thus, we ensure the following for each dataset in the challenge:

\begin{itemize}
\item {\bf Public Code:} All elements of our challenges are on GitHub, providing the full code used for running the challenges through Codabench \cite{Xu2022-lr}.
\item {\bf Common Base Container:} All challenges are built off the same base Docker container with common, standard software tools including \texttt{tensorflow}, \texttt{Pytorch}, and \texttt{NumPy}; it is specified publicly so that the container construction, itself, is reproducible.
\item {\bf Requirements list:} Any changes to the container in terms of packages---either through version upgrades or additional packages---must be specified in a requirements file submitted by participants.
\item {\bf Software Package Whitelist:} Only open-source software packages that the team has reviewed are eligible for installation; in addition to listing the software upgrade in the requirements, these packages are checked against a whitelist of allowed packages to ensure that all software is open-source and thus reproducible. Public GitHub repositories enabled participants to request whitelist additions, while ensuring that these would be documented along with the remaining challenge code.
\end{itemize}

 We place similar conditions on submissions. Thus, requiring they consist of the trained model, with an inference script (containing a model class with at least two specified methods), and a list of software requirements. To illustrate how this is constructed, sample submissions, along with training code and notebooks are provided in sample repositories on GitHub, demonstrating the expected format of the final submissions.
 The challenge repositories are \href{https://github.com/a3d3-institute/HDRchallenge}{Gravitational Waves} (includes the Docker container), \href{https://github.com/Imageomics/HDR-anomaly-challenge}{Hybrid Butterflies}, and \href{https://github.com/iharp-institute/HDR-ML-Challenge-public}{Sea Level Rise}.

For the ML Challenge framework, we choose Codabench~\cite{Xu2022-lr}, a flexible open-source platform designed for benchmarking ML algorithms---particularly through ML challenges---that enables custom code for scoring and presentation of the results. Despite its robust deployment, 
the above listed points are not built into the Codabench framework, requiring us to develop these elements to ensure the challenge was fully FAIR. We thus publish all components of our challenges on GitHub, providing all source code used to run the challenges through Codabench \cite{Xu2022-lr}.
The front end has a standard interface to view scoring, the leaderboard, and instructions for how to participate in the challenge. The inference backend is flexible, so submissions on the Codabench platform are 
run at National Energy Research Scientific Computing (NERSC) Center at Lawrence Berkeley National Laboratory\cite{Bhimji:2024bcd}. Though this adds additional complications in ensuring the submissions are FAIR, it furthers the reproducibility by standardizing the scoring and requiring the whitelist.

Through the use of both
NERSC and the insistence on a FAIR framework, we endeavor to democratize the ML Challenge. Further extensions can be made to
ensure FAIR principles can be adapted to other aspects of the ML challenge, such as model construction, in future editions.
%
%
The final FAIRness component of the challenge is the requirement that all participants publish their challenge submissions (training code, requirements, model weights, etc.) in fully documented GitHub repositories (following the \href{https://github.com/Imageomics/HDR-anomaly-challenge-sample}{Butterfly sample repository} setup). Additionally, when working from pre-trained models, we require that the participants select only those that are also open-source. This ensures that the challenge adheres to a more general notion of fairness in addition to the FAIR principles.


\section{The Challenges}
\label{sec:challenges}

\subsection{Detecting Anomalous Gravitational-Wave Signals}
\label{sec:A3D3-challenge}
Gravitational waves as detected by the large ground-based interferometers LIGO~\cite{LIGOref}, Virgo~\cite{TheVirgo:2014hva} and KAGRA~\cite{KAGRA:2020tym} result from astrophysical phenomena involving the bulk motion of mass at high velocity. They appear as stretching and squeezing of the interferometers' arms due to perturbations of the spacetime metric. So far, all gravitational-wave detections that have been announced correspond to short-duration signals (referred to as transients, or bursts) 
from binary compact mergers involving black holes and neutron stars~\cite{LIGOScientific:2020ibl,LIGOScientific:2021usb,PhysRevX.9.031040,LIGOScientific:2021djp}. These astrophysical systems and their corresponding gravitational-wave emission are well understood and their signatures are modeled so that templated searches (matched filtering) can perform optimal filtering for them in interferometric data. A wealth of astrophysical sources that may potentially emit short-lived gravitational radiation for which we know very little, or close to nothing, on their signal morphologies exist. This includes core-collapse supernovae, neutron star glitches, emission associated with magnetars, other unknown astrophysical systems that power the Fast Radio Bursts, or even topological defects. These transient sources are referred to as unmodeled and they are prime candidates for ML-based anomaly detection approaches.

Gravitational-wave signals are extremely small. A typical gravitational wave on Earth induces a
fractional differential arm change of approximately $10^{-22}$, denoted strain. 
With a strain projected onto the km-scale arms of the interferometers, it results in arm
 displacements thousands of times smaller than a proton’s diameter~\cite{o3performance}.
A single 
interferometer can achieve the sensitivity required to detect gravitational waves. However, it is limited in its capability to distinguish actual signals from large glitches in the detector.
These glitches are often unmodeled transient ``anomalies~\cite{Cabero:2019orq}", denoted glitches that originate from events on Earth such as earthquakes or subtle sources of noise. Multiple detectors are required to veto glitches, which occur in a single detector, as opposed to a signal, which will be in both. The first gravitational-wave event was
observed within the two detectors that constitute the Laser Interferometer Gravitational-wave Observatory (LIGO) in 2015~\cite{LIGOref,NobelRef}. 
Since then, additional gravitational-wave detectors have become operational, allowing for many ways to confirm a signal across multiple instruments and enhancing our ability to identify gravitational-wave events.

The challenge presented here is to find unmodeled astrophysical transients using the two LIGO detectors. To detect such an event, a signal must be observed in both detectors with a time correlation consistent with a wave propagating at the speed of light and a waveform morphology that is correlated across detectors. Since no gravitational-wave event beyond binary black hole and neutron star mergers has been observed, we rely on astrophysical simulations to inject a variety of synthetic signals into the dataset for identification~\cite{targeted_SN_O1-2, O2magnetarbursts, S6_NS, o2_mem,  O3cosmicstring,Robinet:2020lbf,alex_nitz_2020_3993665}. Unmodeled searches have been widely used in the gravitational-wave community and reported in observational papers~\cite{Klimenko:2008fu,PhysRevD.95.104046,Macquet:2021ibe,Abbott_2021, KAGRA:2021bhs}.
Furthermore, recent efforts to perform AI-based anomaly detection have emerged\cite{verma2024detectiongravitationalwavesignals, PhysRevD.103.102003, Krastev_2020, PhysRevD.97.044039, PhysRevD.108.024022,Skliris:2020qax,que2021accelerating, Raikman:2023ktu, raikman2024neuralnetworkbasedsearchunmodeled}; this is the primary focus of the challenge presented here.

The Anomaly Detection Challenge utilizes data from LIGO's O3a observing run, consisting of 
calibrated strain time-series recorded by the two LIGO interferometers.  The provided dataset has undergone a series of pre-processing steps: whitening, band-passing, and removing from the dataset altogether 1s worth of data around the times of gravitational-wave detections made and published by the LIGO-Virgo-KAGRA collaborations.
Additionally, the datasets contain simulated signals injected into the real detector noise, as described in Section~\ref{sec:GW_data}. 
Participants are required to train their models for anomaly detection primarily using the background data and can improve their ability to identify anomalies through injected simulated signals. 

\subsubsection{Data samples}
\label{sec:GW_data}
\begin{figure}[tb]
\centering
\includegraphics[width=0.6\textwidth]{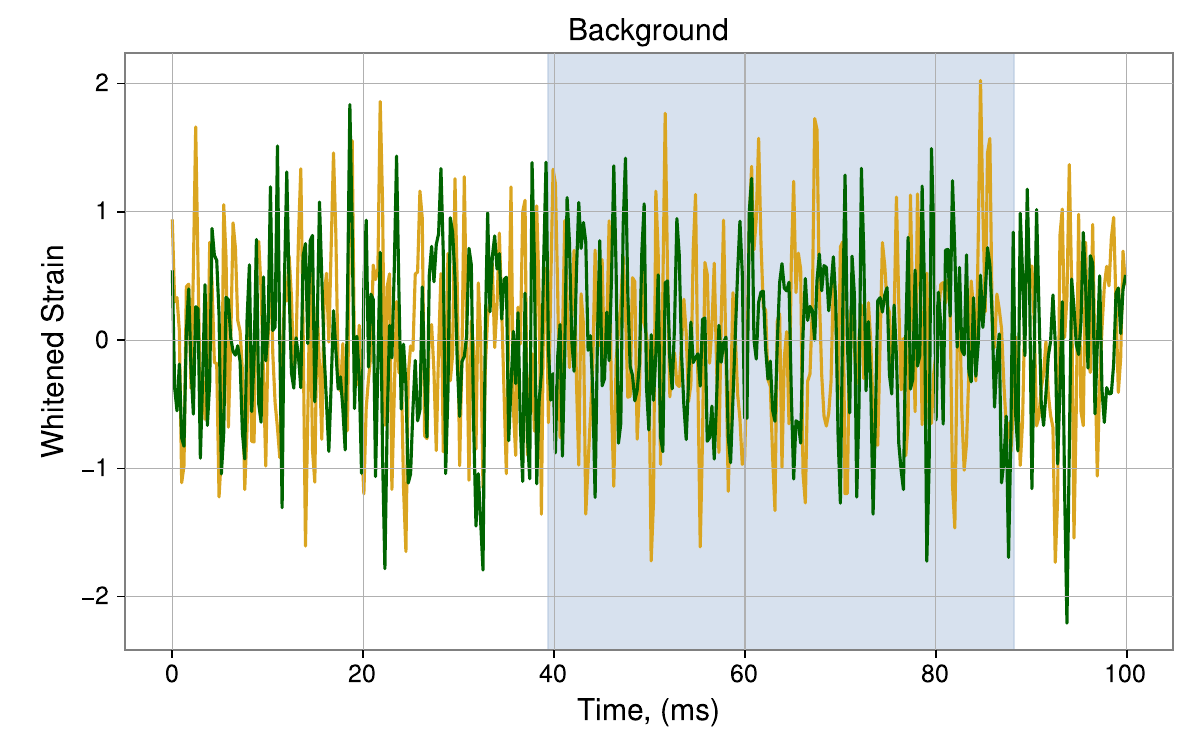}
\caption{Example of the background strain. The light blue
shading highlights an example region that is passed as input to the autoencoders for training. The yellow and green lines indicate the strain from the two LIGO detectors Hanford~(yellow) and Livingston~(green). }
\label{fig:background_classes}
\end{figure}

\begin{figure}[tb]
\centering
\includegraphics[width=0.6\textwidth]{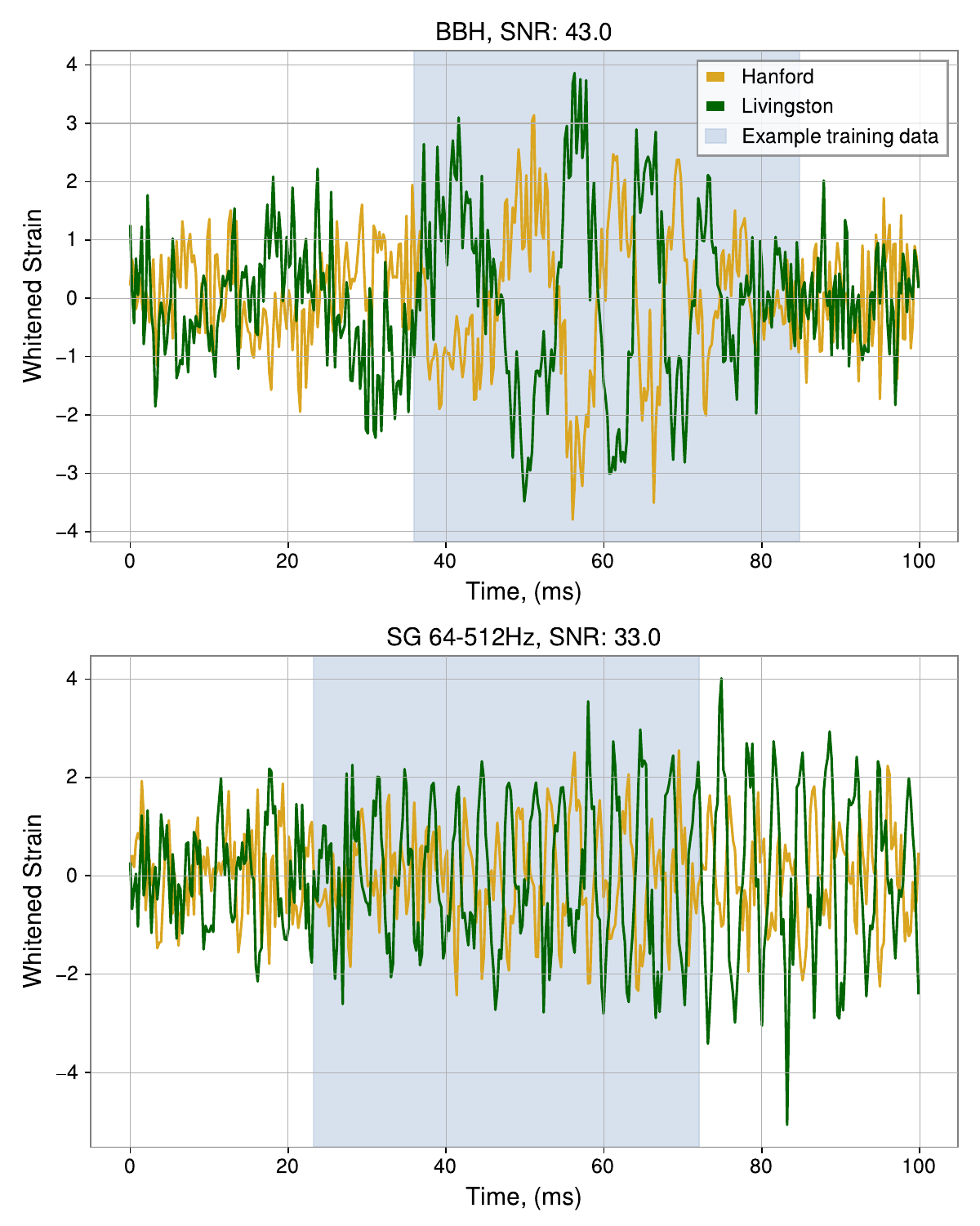}
\caption{Example of signal-like classes: Binary Black Hole Mergers (top) and sine-gaussian (bottom) strains. The light blue shading highlights an example region that is passed as input to the autoencoders for training.}
\label{fig:signal_classes}
\end{figure}

The dataset used in this challenge was collected by the LIGO Hanford and LIGO Livingston~\cite{TheLIGOScientific:2014jea}. 
We utilized publicly available data from the beginning of observing run O3a~\cite{LOSCref}, corresponding to GPS times 1238166018 to 1238170289. 
The time-series data were 
sampled at 4096\,Hz, and processed to remove and create a separate dataset of transient instrumental artifacts (glitches).
We extracted 4\,s segments of artifact-free data to serve as the baseline for the injection of signals.
We define
three data classes representing signals and background signatures:
\begin{itemize}
    \item \textbf{Binaries} -- Simulated Binary Black Holes~(BBH) signals injected into the real background noise, as shown in Fig.~\ref{fig:signal_classes} (top). 
    \item \textbf{Background} -- Background from O3a with excess power glitches and known gravitational-wave events removed, 
    as shown in Fig.~\ref{fig:background_classes}.
    \item \textbf{Sine-Gaussian (SG)} -- Generic ad-hoc
    signal model used as a proxy to simulate gravitational-wave sources we know little about, as shown in Fig.~\ref{fig:signal_classes} (bottom). 
\end{itemize}
To generate samples of BBH and SG signals, we perform simulations of the two polarization modes, $h_+$ and $h_\times$, which describe the strain-induced distortions in spacetime.
We then sample sky localizations uniformly in the sky, 
generate the polarization modes 
at the specified sky location, and then inject the projected modes into the two LIGO detectors.
\noindent

We whiten the data and band-pass them within the frequency range of 30--1500\,Hz.
After applying these filters, we remove 1\,s intervals from each end of the data samples to eliminate edge effects from pre-processing. The remaining 2\,s samples, each containing either an injected signal, pure background or an SG, were used to generate training data.
To obtain a set of windows suitable for training, we extract 200 data points (total duration of 50\,ms sampled at 4096\,Hz) from each sample. 

\subsubsection{Challenge}
The challenge is to identify astrophysical signals that are neither background noise nor glitches within the detector. 
Astrophysical signatures have the characteristic features shown in Figure ~\ref{fig:signal_classes} whereby the two detectors yield inverted signals with roughly the same amplitude. These signals are time-aligned such that the arrival of the signals corresponds to the same gravitational wave propagating from a given location in the sky. The sky location will change the relative amplitudes, along with the knowledge of the noise levels at each of the respective detectors making it possible for an asymmetry in amplitudes between the detectors. The signals themselves are limited to frequency ranges from roughly 10 Hz to 1000 Hz.


\subsection{ Detecting Hybrid Butterflies}

The criteria for elevating subspecies to species remain an open question in evolutionary biology. More broadly, these taxonomic distinctions have significant implications for conservation efforts, as they influence which populations receive legal protection under the Endangered Species Act\cite{Zink2022-vc}. A key challenge in resolving species boundaries is our ability to detect and classify subspecies, particularly in cases where hybridization occurs. The question of how hybridization contributes to the evolution and maintenance of new species has intrigued scientists since Charles Darwin, who pondered the variability and stability of hybrid traits. Gregor Mendel later addressed this question through his experiments, demonstrating that hybrid offspring do not always exhibit intermediate traits, instead resembling one parent, following predictable patterns of segregation \cite{mendel1865verhandlungen}. 

These principles of inheritance remain central to modern hybrid detection, particularly in distinguishing whether hybrids between subspecies exhibit continuous variation or Mendelian inheritance patterns. This challenge is designed to leverage anomaly detection algorithms to detect hybrids produced by parents of hybridizing subspecies. By refining hybrid detection methods, we can improve taxonomic classification and enhance conservation strategies for threatened populations.


\begin{figure}[!htbp]
\centering
\includegraphics[width = \linewidth]{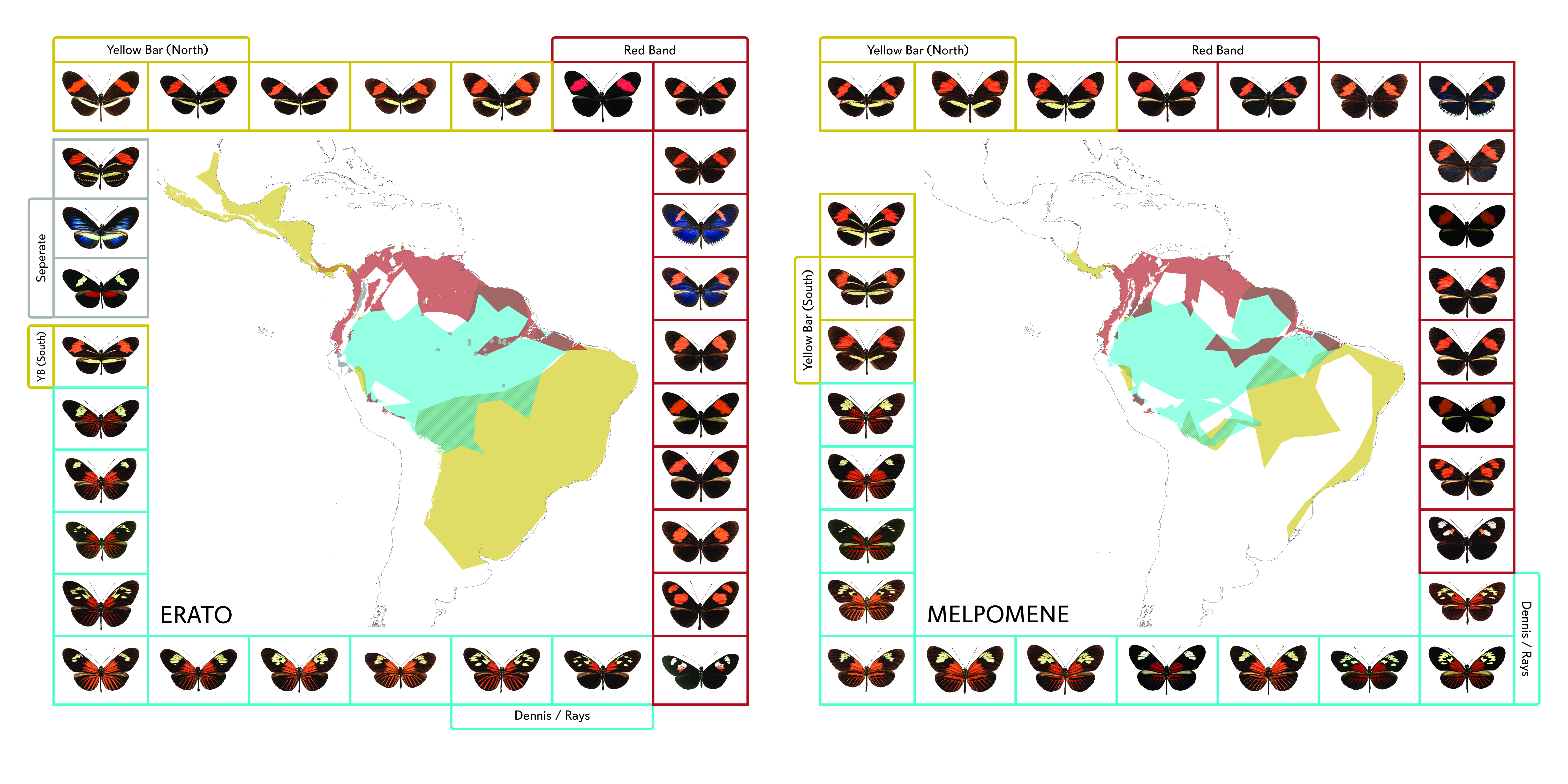}
\caption{\small The parallel radiation of mimetic color-pattern complexes in \textit{Heliconius erato} (left) and \textit{Heliconius melpomene} (right). Each region demarcates the boundaries of a different color pattern.
See~\cite{hines2011wing} for more details. 
}
\label{fig:butterfly-range}
\end{figure}

\subsubsection{Background}\label{butterfly-background}
\textit{H. erato} and \textit{H. melpomene} are two \textbf{species} of Heliconius butterflies (Order: Lepidoptera; Family: Nymphalidae) that diverged approximately 12 million years ago. Both species are chemically defended (unpalatable) and signal their toxicity through bright warning coloration on their wings. Interestingly, they have evolved to resemble each other, forming a classic example of M\"{u}llerian mimicry, in which two or more toxic species share similar warning signals to reinforce predator learning \cite{mallet2007}. This is distinct from \textbf{Batesian} mimicry, where a palatable species mimics an unpalatable one to gain protection. Despite strong predation pressure to maintain similar warning patterns, \textit{Heliconius} species and their color morphs have diversified across Central and South America in what biologists term \textit{adaptive radiation}. That is, distinct geographic regions---sometimes microhabitats within these regions---harbor unique assemblages of \textit{H. erato} and \textit{H. melpomene}, each displaying region-specific wing color patterns that serve as distinguishing characteristics. 




The \emph{visually different appearances among subspecies in different regions} and \emph{visually mimicking appearances between species in the same regions} result in large intra-species variation within \textit{H. melpomene} (\textit{H. erato}) and small inter-species variation between \textit{H. melpomene} and \textit{H. erato} (Figure~\ref{fig:butterfly-range}).
This phenomenon has attracted attention in biology, ecology, computer vision, and machine learning to study the visual traits distinguishing subspecies and species of Heliconius butterflies from each other.

\subsubsection{Anomaly cases}

\textit{Heliconius} butterflies exhibit strong assortative mating, preferentially choosing mates with the same wing color pattern. This behavior is reinforced by natural selection---\textbf{hybrids} resulting from crosses between individuals with different patterns often face a survival disadvantage. Their unfamiliar wing patterns are not recognized by local predators, increasing their likelihood of being removed from the population. However, where subspecies come into contact or overlap, interbreeding still occurs, producing hybrids with diverse phenotypes. The visual appearance of hybrids can vary—some resemble one parent more, while others exhibit an intermediate or novel pattern. Historically, these hybrid forms were labeled as aberrations or given distinct “form names.” If such hybrids become frequent enough to evade predation, they may persist in the population and potentially contribute to speciation. It is these hybrids, aberrations, and forms that we aim to detect.

\subsubsection{Scenario}
\label{sec:butterfly-scenario}
We design our competition to simulate a real-world biological scenario. Suppose a biologist studies \textit{Heliconius melpomene} and \textit{Heliconius erato}, aiming to understand the mimicry phenomenon and the different color patterns of subspecies. One day, she finds that a subset of the butterfly collections looks slightly abnormal in their color patterns. After further investigation, she discovers these samples are hybrids generated by different, rarely observed, subspecies of 
\textit{H. erato}. She realizes there may also be hybrids among the \textit{H. melpomene} specimens, but has fewer samples. Since in theory, there are quadratically many cases of hybrids and her current collection only covers a small subset of them, she seeks an anomaly detection system to automatically identify (unknown) hybrid cases from future collections of Heliconius butterflies---both \textit{H. erato} and \textit{H. melpomene}.

\subsubsection{Data and Splits}

\textbf{Data}

The dataset is comprised of a subset of the Heliconius Collection (Cambridge Butterfly) \cite{lawrence_campolongo_j2024}, a compilation of images from Chris Jiggins' research group at the University of Cambridge \cite{jiggins_2019_2549524, jiggins_2019_2550097, jiggins_2019_2682458, mattila_2019_2555086, joana_i_meier_2020_4153502, montejo_kovacevich_2019_2677821, montejo_kovacevich_2019_2684906, montejo_kovacevich_2019_2686762, montejo_kovacevich_2019_2702457, montejo_kovacevich_2019_2707828, montejo_kovacevich_2019_2714333, montejo_kovacevich_2019_2813153, montejo_kovacevich_2019_3082688, montejo_kovacevich_2021_5526257, 
gabriela_montejo_kovacevich_2020_4289223, gabriela_montejo_kovacevich_2020_4291095, patricio_a_salazar_2020_4288311, pinheiro_de_castro_2022_5561246, salazar_2018_1748277, salazar_2019_2548678, salazar_2019_2735056, warren_2019_2552371, warren_2019_2553501, warren_2019_2553977}.
%
It encompasses two aspects of biological development and evolutionary change: (1) hybridization---the main theme of this challenge---and (2) mimicry.


\begin{enumerate}
    \item \textbf{Hybridization:}
    
    Geographic or habitat separation of a species population can lead to species variation developing into different subspecies. The visual appearances (e.g., color patterns on the wings) of these subspecies can be drastically different.
%
%
%
In this challenge, offspring produced by the same-subspecies parents (\textbf{non-hybrids}) are treated as normal cases because they are far more frequently observed. In contrast, \textbf{hybrids} are treated as \textbf{anomaly} cases, not only because they are much less frequently observed---with some combinations not yet observed---but also because their visual appearances are much more variant and hardly predictive.
We emphasize again that the parents of each hybrid child are from different subspecies of the \textit{same} species (either \textit{H. melpomene} or \textit{H. erato}).

\item \textbf{Butterfly mimicry}

Meanwhile, geographic or habitat overlap can also lead to increased visual similarity between species, known as \textit{mimicry}. 
This challenge goes beyond developing an anomaly detection algorithm to distinguish between hybrids and non-hybrids for one species, investigating further whether such an algorithm is generalizable to the visually mimetic species.

\end{enumerate}



\medskip

\noindent{}\textbf{Splits}

\begin{figure}[!htb]
\centering
\includegraphics[width=0.7\textwidth]{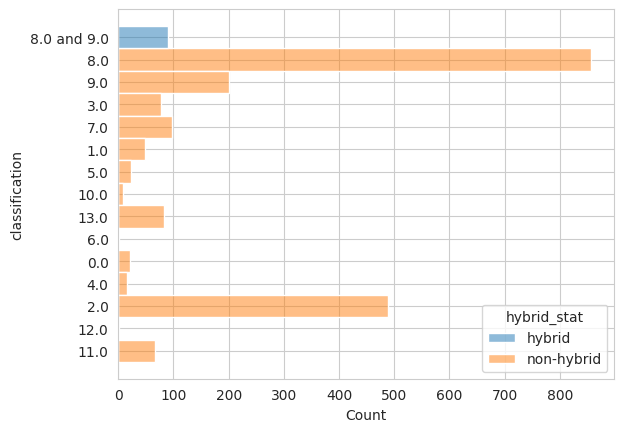}
\caption{Distribution of anonymized \textit{H. erato} subspecies included in the training data. This figure was generated in a starter Jupyter Noteboook provided to participants in the \href{https://github.com/Imageomics/HDR-anomaly-challenge-sample}{sample repository}.
}
\label{fig:butterfly_training_dist}
\end{figure}

The training data comprises images from all the 
\textit{Heliconius erato} subspecies and the \textbf{signal hybrid}: a specific combination of the parent subspecies that has the most images (these hybrids are the most common \emph{within this particular dataset, not necessarily in general}; see Figure~\ref{fig:butterfly_training_dist}). All other hybrids, \textbf{non-signal hybrids}, are excluded from the training data.

\begin{figure}[!htb]
\centering
\includegraphics[width=0.9\textwidth]{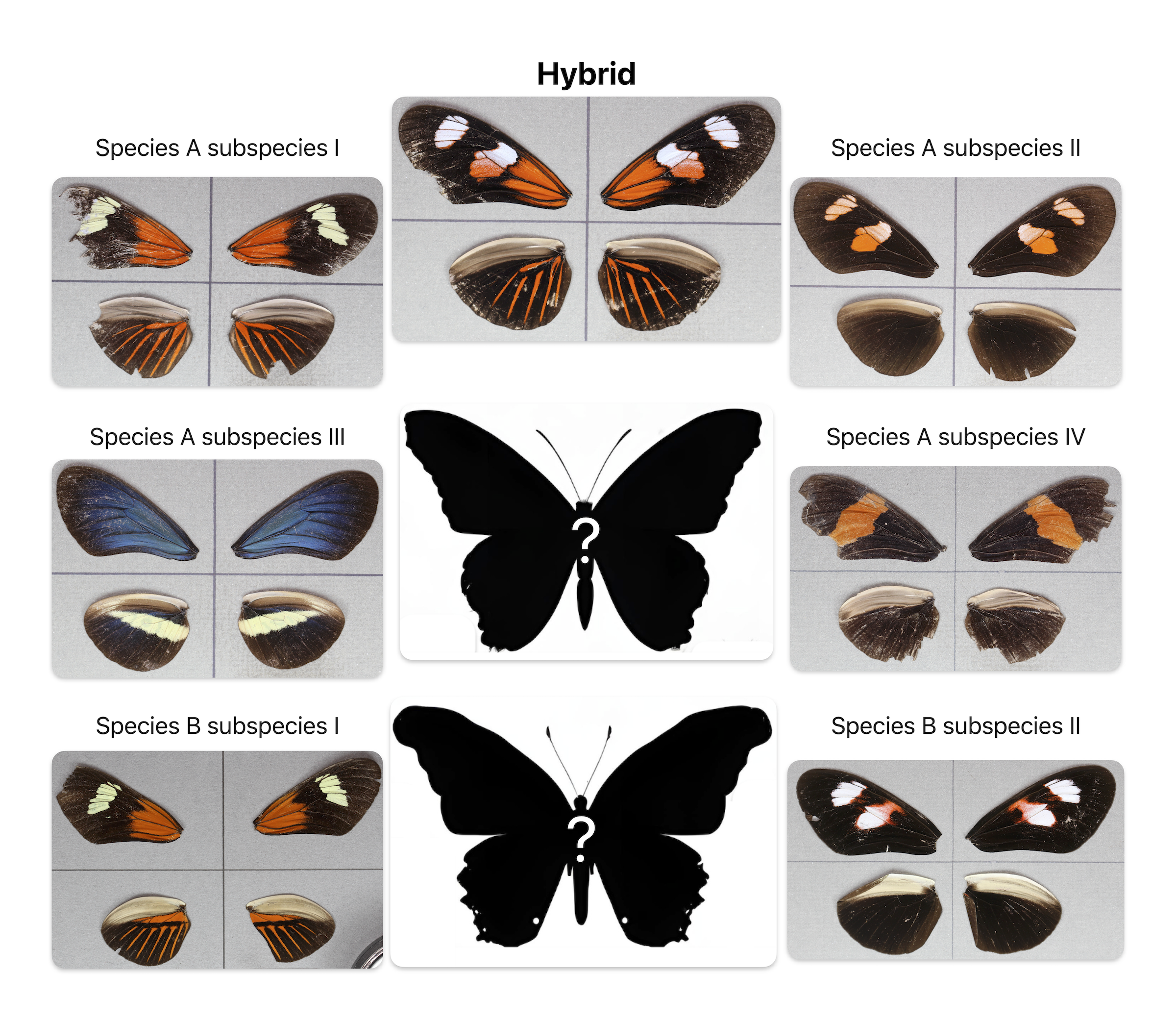}
\caption{
For the challenge, participants were only provided the labels ``Species A subspecies \#".
These four subspecies of Species A (\textit{H. erato}) are seen in training, along with the signal hybrid (hybrid of \textit{Species A subspecies I \& II}), but can we detect the hybrid of \textit{Species A subspecies III \& IV} when we do not know what they look like? Additionally, can we distinguish the Species B (\textit{H. melpomene}) subspecies I \& II hybrids, considering these subspecies mimic \textit{Species A subspecies I \& II}, respectively. Images are from \cite{jiggins_2019_2549524, montejo_kovacevich_2019_2677821, montejo_kovacevich_2019_2686762, montejo_kovacevich_2019_2714333, montejo_kovacevich_2019_3082688}.
The hybrid graphics (black butterflies) were generated using Canva Magic Media AI, then manually edited.}
\label{fig:butterfly_hybrids_Q}
\end{figure}

We consider the following two sets of images in the test set, as illustrated in Figure~\ref{fig:butterfly_hybrids_Q}.
\begin{enumerate}
    \item All Species A (\textit{Heliconius erato}) subspecies and their hybrids, including the signal hybrid and the non-signal hybrids.


\item Two subspecies of Species B (\textit{H. melpomene}) and their hybrid. 
%
These subspecies of
\textit{H. melpomene} are those that mimic the parent subspecies of the signal hybrid of \textit{H. erato}. 
\end{enumerate}


\subsection{Detecting Anomalous Sea Level Rise Events}
\label{iHARP-challenge}

The US East Coast is a region particularly vulnerable to coastal flooding and storm surges. Daily tide gauge data from the National Data Buoy Center (NDBC) \cite{ndbc} provide essential observations, including tidal variations, storm surges, and long-term trends, forming the foundation for this challenge. 
Additionally, long-term sea level records are invaluable for studying climate change impacts, providing evidence of global warming effects like melting polar ice and thermal expansion of seawater. These observations support the resilience and sustainability of coastal economies, which rely heavily on tourism, fishing, and shipping industries. 
Predicting sea level anomaly events, such as extreme storm surges or unusually high tides, is challenging along the low-lying US East Coast region due to the complex interplay of atmospheric, oceanic, and climatic factors. These events are influenced by a combination of wind patterns, atmospheric pressure changes, and ocean currents, making accurate forecasting difficult with traditional methods.  By leveraging ML to process real-time data from NDBC buoys, historical sea level records, and meteorological information, participants are tasked with detecting and predicting anomalous sea-level rise events caused by factors such as hurricanes, mid-latitude storms, or long-term climatic phenomena like the El Niño Southern Oscillation (ENSO) \cite{enso}. If accurately detected, these anomalies can be vital in enhancing coastal community preparedness and safeguarding lives, infrastructure, and economic activities.

\begin{figure}[tb]
\centering

\includegraphics[width=0.8\textwidth]{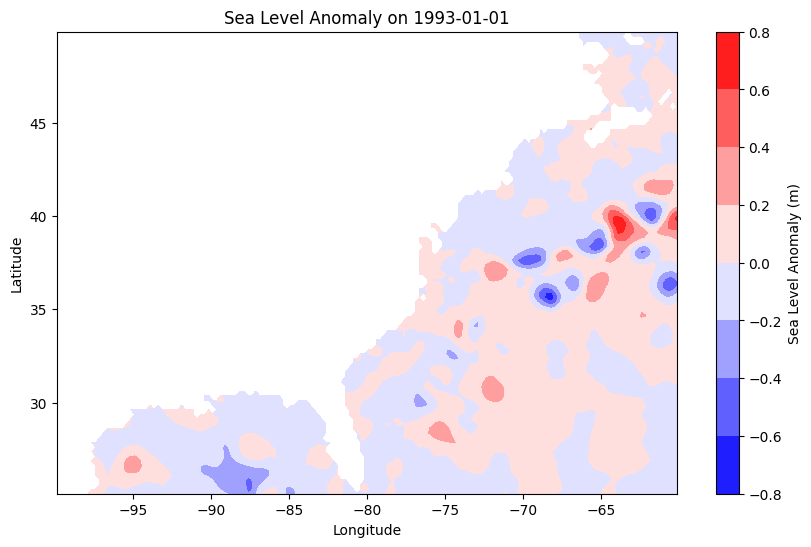}
\caption{Depiction of the sea levels on the Eastern Seaboard from January 1st, 1993 obtained from satellite images of the US eastern seaboard.}
\label{fig:sea-level rise}
\end{figure}

\subsubsection{Objective}
The objective is to predict anomalous sea-level observations from daily tide gauge data along the US East Coast affected by changes in the sea-level elevation values on the Atlantic Ocean. The challenge leverages a comprehensive training dataset that spans 20 years of daily sea-level measurements from 12 coastal stations along the US East Coast, complemented by regional satellite sea-level elevation data in the North Atlantic. The satellite images are referred to as the Copernicus dataset~\cite{CDS_portfolio}, which offers a broader spatial context, enabling participants to incorporate regional oceanic conditions into their models. Therefore, participants are required to develop models that take the satellite maps of spatial information of sea level anomalies over the North Atlantic as input and predict dates when coastal stations along the US East Coast are above a flooding threshold for the subsequent period of 10 years, which is the test dataset. Each submission must specify whether a flooding level anomaly occurred daily at each station.
This task tests the participants’ ability to build robust predictive models and emphasizes the importance of precision and recall in capturing true anomalies.

\subsubsection{Evaluation}
Evaluation metrics, including the average true positive rate, average false positive rate, and the F1 score, guide the assessment of model performance. 
The challenge fosters innovation, collaboration, and the development of scalable solutions with direct societal benefits, ultimately advancing our ability to predict and respond to sea-level anomalies, thereby strengthening the resilience of vulnerable coastal regions.


\subsubsection{Datasets}
The core dataset includes hourly sea-level measurements from 12 tide gauge stations from 1993 to 2013 for each station. Gridded Sea Level Anomalies (SLA) computed for a 20-year mean are also provided. These SLA values are estimated using Optimal Interpolation, merging along-track measurements from multiple altimeter missions processed by the DUACS multimission altimeter data system. Additional variables, such as Absolute Dynamic Topography and geostrophic currents (both absolute and anomalous), are also included, making the dataset suitable for delayed-time applications and allowing participants to explore the dynamic relationships between sea-level changes and broader oceanic processes.

Each coastal station’s data is represented as a distinct time series, and a separate column identifies the dates associated with known anomalies. This dataset captures critical sea-level fluctuations caused by various climatic and oceanic phenomena \cite{ghosh2024towardssigspatial}. Additionally, the Copernicus sea-level gridded satellite observations dataset \cite{CDS_portfolio} contributes sea-level elevation values from the Atlantic Ocean for the same period, offering a broader contextual dataset to enhance the predictive capabilities of the models.

The training dataset is a labeled subset of the time series data from the tide gauge stations. This dataset includes both the raw sea-level measurements and the associated anomaly labels. 
The combination of labeled anomalies and continuous measurements allows for the exploration of various ML approaches, such as supervised learning, time series analysis, and feature engineering.

The public testing dataset includes a subset of the time series data from each station but does not contain labeled anomalies. This dataset allows participants to 
refine their methodologies. The challenge dataset, on the other hand, includes hidden anomalies and will be used to evaluate the final performance of submitted models. 

By offering labeled and unlabeled data across diverse coastal stations, this challenge emphasizes the development of models that can generalize well to unseen data and effectively identify anomalies in complex, multivariate time series. Following FAIR principles ensures that the solutions generated will have practical applications in predicting and responding to coastal sea-level anomalies, ultimately contributing to improved resilience and preparedness in vulnerable regions.

\section{Metrics}
\label{sec:metrics}
\paragraph{The evaluation metric is the false positive rate (FPR) at a specified true positive rate (TPR) when detecting anomalies.} The specified TPR, which we denote by $\text{TP}\%$, varies by challenge, as indicated in Table~\ref{tab:TPR}. 
Given a test set with both normal and anomalous
signals/instances,
this metric treats the normal cases as positive cases and finds a score threshold $\tau$ such that $\text{TP}\%$ of the normal 
signals have scores $s(\textit{\textbf{x}})\geq\tau$. It then counts the percentage of
anomalous signals whose score $s(\textit{\textbf{x}})\geq\tau$ (i.e., FPR). The higher the FPR is, the poorer the anomaly detection algorithm performs.

\begin{table}[!h]
    \centering
    \begin{tabular}{|l|c|}
    \hline
       \textbf{Dataset}  &  \textbf{True Positive Rate Threshold} \\
       \hline
       \hline
       Gravitational-wave signals  & 90\%\\
       \hline
       Hybrid butterflies  & 95\% \\
       \hline
       Sea level rise  & 95\% \\
       \hline
    \end{tabular}
    \caption{True Positive Rate (TPR) threshold chosen for each dataset to calculate the challenge scores.}
    \label{tab:TPR}
\end{table}


\section{Conclusion}
\label{sec:conclusions}
In summary, we have developed three scientific datasets targeting anomaly detection and deployed machine learning challenges. These datasets provide concrete examples of how anomaly detection plays an important role in scientific discovery. Moreover, the implications of a successful model within any of these challenges would be direct in the scientific domain, and---in some cases---profound.  In preparing these challenges, we have emphasized reproducibility partnered with the use of \textbf{F}indable \textbf{A}ccessible \textbf{I}nteroperable and \textbf{R}eusable principles. 
Our current challenge has recently completed, with more than 600 participating teams spread over the three different datasets. 

The resulting solutions can have broad impacts in many domains. Time series anomalies within the gravitational-wave problem have direct implications on anomalous neural activity, such as sleep spindles\cite{TapiaRivas2024ARD}. Equivalently, the hybrid detection problem has direct implications for processing medical images and diagnoses\cite{carloni2022applicability, chen2019looks}. Lastly, the climate science problem can be applied more broadly to satellite data to predict future catastrophic effects\cite{yang2013role, YANG2025102019}. 

When considering future science-based challenges, it is critical to ensure a public codebase that includes the full scoring and example submissions is available. The use of a computing backend with sufficient GPUs has been essential. Though raising the complexity, it increases reproducibility. Going forward, we recommend that future challenges consider these essential elements to expedite the path from a challenging problem to scientific discovery.


\bibliography{ref.bib,Harry.bib}

\end{document}